\documentclass[conference]{IEEEtran}

\newcommand{\gbf}[1] {\mbox{\boldmath${#1}$\unboldmath}}
\newcommand{\be}{\begin{equation}}
\newcommand{\ee}{\end{equation}}
\newcommand{\beq}{\begin{equation}}
\newcommand{\eeq}{\end{equation}}
\newcommand{\bed}{\begin{displaymath}}
\newcommand{\eed}{\end{displaymath}}
\newcommand{\beqa}{\begin{eqnarray}}
\newcommand{\eeqa}{\end{eqnarray}}
\newcommand{\beqann}{\begin{eqnarray*}}
\newcommand{\eeqann}{\end{eqnarray*}}
\newcommand{\bseq}{\begin{subequation}}
\newcommand{\eseq}{\end{subequation}}

\newcommand{\ba}{\begin{array}}
\newcommand{\ea}{\end{array}}

\newcommand{\negr}[1]{{\bf {#1}}}

\usepackage{epsfig}
\usepackage{subeqn}
\usepackage{subfigure}

\begin{document}

%%%%% TITLE %%%%%%%%%%%%%%%%%%%%%%%%%%%%%%%%%%%%%%%%%%%%%%%%%%%%%%%%%%%%%%%%%%%%
\title{Working and Assembly Modes of the Agile Eye}
\author{\begin{tabular}[t]{cc}
{  Ilian A. Bonev} & 
{Damien Chablat and  Philippe Wenger}\\
	D\'epartement de g\'enie de la production automatis\'ee &
  Institut de Recherche en Communications \\
  \'Ecole de Technologie Sup\'erieure &
  et Cybern\'etique de Nantes, \\
 1100 Notre-Dame St. West, Montr\'eal, QC, Canada H3C 1K3 &
 1 rue de la No\"e, 44321 Nantes, France \\
 \small{Ilian.Bonev\symbol{64}etsmtl.ca} &
 \small{\{Damien.Chablat, Philippe.Wenger\}@irccyn.ec-nantes.fr} \\
  \end{tabular}}

\maketitle
\pagestyle{empty}
\thispagestyle{empty}

%%%%% ABSTRACT %%%%%%%%%%%%%%%%%%%%%%%%%%%%%%%%%%%%%%%%%%%%%%%%%%%%%%%%%%%%%%%%%
\section*{Abstract}
This paper deals with the in-depth kinematic analysis of a special spherical parallel wrist, called the \textit{Agile Eye}. The \textit{Agile Eye} is a three-legged spherical parallel robot with revolute joints in which all pairs of adjacent joint axes are orthogonal. Its most peculiar feature, demonstrated in this paper for the first time, is that its (orientation) workspace is unlimited and flawed only by six singularity curves (rather than surfaces). Furthermore, these curves correspond to self-motions of the mobile platform. This paper also demonstrates that, unlike for any other such complex spatial robots, the four solutions to the direct kinematics of the \textit{Agile Eye} (assembly modes) have a simple geometric relationship with the eight solutions to the inverse kinematics (working modes).

%%%%% SECTION 1 %%%%%%%%%%%%%%%%%%%%%%%%%%%%%%%%%%%%%%%%%%%%%%%%%%%%%%%%%%%%%%%%
\section{Introduction}

Most of the active research work carried out in the field of parallel robots has been focused on a particularly challenging problem, namely, solving the direct kinematics, that is to say, finding each possible position and orientation of the mobile platform (\textit{assembly mode}) as a function of the active-joint variables. A second popular problem has been the evaluation and optimization of the workspace of parallel robots \cite{Gosselin:1988,Merlet:2000}. Unfortunately, the direct kinematic problem and the workspace analysis have most often been treated independently, although they are closely related to each other. 

It is well known that most parallel robots have singularities in their workspace where stiffness is lost \cite{Gosselin:1990}. These singularities (called {\em parallel singularities} in \cite{Chablat:1998} and Type 2 in \cite{Gosselin:1990}) coincide with the set of configurations in the workspace where the finite number of different direct kinematic solutions (assembly modes) changes. For parallel manipulators with several solutions to the inverse kinematic problem (\textit{working modes}), another type of singularity exist and defines what may be generalized as the workspace boundary. These singularities (called {\em serial singularities} in \cite{Chablat:1998} and Type 1 in \cite{Gosselin:1990}) coincide with the set of configurations in the workspace where the finite number of different inverse kinematic solutions (working modes) changes. While ensuring that a parallel robot stays with the same working mode along a discrete trajectory is straightforward, the notion of ``same assembly mode'' is not even clear in the general case. Indeed, the direct kinematic solutions are most often obtained by solving a univariate polynomial of degree $n>3$, which means that there is no way to designate each solution to a particular assembly mode. Then, how does one choose a direct kinematic solution for a parallel robot for each different set of active-joint variables?

A natural sorting criterion is that the direct kinematic solution could be reached through continuous motion from the initial assembly mode, i.e., the reference configuration of the robot when it was first assembled, without crossing a singularity. Before \cite{Innocenti:1992}, it was commonly thought that such a criterion was sufficient to determine a unique solution. Unfortunately, as for serial robots \cite{Innocenti:1992,Wenger:1992}, a non-singular configuration changing trajectory may exist between two assembly modes for robots that are called \textit{cuspidal}. This result later gave rise to a theoretical work on the concepts of characteristic surfaces and uniqueness domains in the workspace \cite{Wenger:1997}. However, it still remains unknown what design parameters make a given parallel robot cuspidal, and it is still unclear how to make a given parallel robot work in the same assembly mode. It will be shown in this paper, that there is a spherical parallel robot (quite possibly the only one), for which there are clear answers to these complex questions.

From the family of parallel wrists \cite{Karouia:2003}, the {\em Agile Eye} provides high stiffness and is quite probably the only one to provide a theoretically \textit{unlimited and undivided orientation workspace}. The {\em Agile Eye} is a 3-\textit{\underline{R}RR} spherical parallel mechanism in which the axes of all pairs of adjacent joints are orthogonal. Based on the {\em Agile Eye} design, a camera-orienting device was constructed at Laval University a decade ago \cite{Gosselin:1996}, an orientable machine worktable was manufactured at Tianjin University, and a wrist for a 6-DOF robot was built at McGill University.

The \textit{Agile Eye} has been extensively analyzed in literature, but surprisingly some of its most interesting features have not been noticed. One of the key references in this paper is \cite{Gosselin:1995}, where the simple solution to the direct kinematics of the \textit{Agile Eye} is presented. Namely, it is shown that the \textit{Agile Eye} has always four trivial solutions (at which all three legs are at singularity and can freely rotate) and four nontrivial solutions obtained in cascade. A second key reference is \cite{Gosselin:2002}, where the singularities of the \textit{Agile Eye} are studied. Unfortunately, in \cite{Gosselin:2002}, it was mistakenly assumed that the only singular configurations are the four trivial solutions (orientations) to the direct kinematics. Basically, the fact that at each of these four orientations, a special arrangement of the legs can let the mobile platform freely rotate about an axis was overlooked. Indeed, as this paper shows, the singularities of the \textit{Agile Eye} are six curves in the orientation workspace corresponding to self motions of the mobile platform or lockup configurations.

When it was realized that the unlimited workspace of the \textit{Agile Eye} is not divided by singularity surfaces (as is the case for all other spherical parallel robots), yet accommodating four unique assembly modes, it first seemed that the \textit{Agile Eye} is a cuspidal robot. In this paper, it is shows not only that the \textit{Agile Eye} is not cuspidal, but that it is straightforward to identify its four assembly modes via a simple relationship to its working modes. The results of this paper provide more insight into the kinematics of the \textit{Agile Eye} and shed some light to the open problem of assembly mode designation.

The rest of the paper is organized as follows. In the next two sections, the kinematic model of the \textit{Agile Eye} and its inverse kinematics are presented briefly. Then, the direct kinematics and the singularity analysis of the \textit{Agile Eye} are addressed by partially reproducing, reformulating, and augmenting the results of \cite{Gosselin:1995} and \cite{Gosselin:2002}. Finally, it is shown that for each working mode there is a single corresponding assembly mode from among the four nontrivial ones.

%%%%% SECTION 2 %%%%%%%%%%%%%%%%%%%%%%%%%%%%%%%%%%%%%%%%%%%%%%%%%%%%%%%%%%%%%%%%
\section{Kinematic Model of the Agile Eye}

%%%%% FIGURE 1 %%%%%%%%---------------------------------------------------------
\begin{figure}[t]
  \begin{center}
    \scalebox{0.25}{\includegraphics{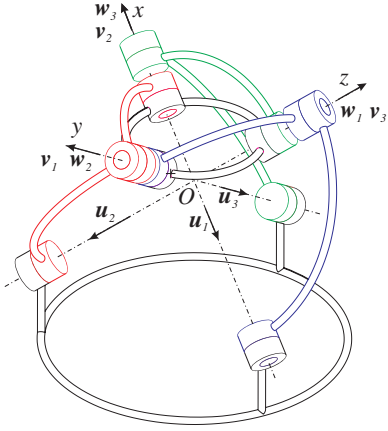}}
    \vspace{-2mm}
    \caption{The \textit{Agile Eye} at its reference configuration.}
    \protect\label{fig:Agile_eye}
  \end{center}
\end{figure}
%%%%%%%%%%%%%%%%%%%%%%%---------------------------------------------------------

Figure~\ref{fig:Agile_eye} depicts the {\em Agile Eye} at its reference configuration, where the mobile platform is at its reference orientation and all active-joint variables are zero. At the reference configuration, the axes of the first base joint, the second intermediate joint, and the third platform joint coincide, and so on for the other joints (the figure is self-explanatory).

A base reference frame $Oxyz$ is chosen in such a way that its $x$ axis is along the axis of the first base joint, its $y$ axis is along the axis of the second base joint, and its $z$ axis is along the axis of the third base joint, as shown in Fig.~\ref{fig:Agile_eye}. A mobile reference frame (not shown) is fixed at the mobile platform so that it coincides with the base frame at the reference orientation. With these settings, the axes of the base joints are defined by the following unit vectors expressed in the base reference frame:
\begin{equation}
  \negr u_1 = \left[ \begin{array}{c} 1 \\ 0 \\ 0 \end{array} \right]\!, \quad
  \negr u_2 = \left[ \begin{array}{c} 0 \\ 1 \\ 0 \end{array} \right]\!, \quad
  \negr u_3 = \left[ \begin{array}{c} 0 \\ 0 \\ 1 \end{array} \right]\!.
\end{equation}

\noindent
Similarly, the axes of the platform joints are defined by the following unit vectors expressed in the mobile reference frame:
\begin{equation}  
  \negr v'_1 = \left[ \begin{array}{c} 0 \\ -1 \\ 0 \end{array} \right]\!, {\rm ~}
  \negr v'_2 = \left[ \begin{array}{c} 0 \\ 0 \\ -1 \end{array} \right]\!, {\rm ~}
  \negr v'_3 = \left[ \begin{array}{c}-1 \\ 0 \\ 0  \end{array} \right]\!.
\end{equation}

The rotation matrix \negr R describes the orientation of the mobile frame with respect to the base frame. The \textit{ZYX} Euler-angle convention is used here because it simplifies greatly the kinematic analysis. In the base frame, the axes of the platform joints are defined as follows
\begin{equation}
   \negr v_i = \negr R \negr v'_i = \negr R_z(\phi)\negr R_y(\theta)\negr R_x(\psi) \negr v'_i,
\end{equation}
where $i = 1, 2, 3$, which yields
\begin{eqnarray}
  \negr v_1&\!\!\!\!=\!\!\!\!&
  \left[
    \begin{array}{c}
    \sin\phi \cos\psi -\cos\phi \sin\theta \sin\psi \\
   -\cos\phi \cos\psi -\sin\phi \sin\theta \sin\psi \\
   -\cos\theta \sin\psi 
    \end{array}
  \right]\!, \nonumber \\
  \negr v_2&\!\!\!\!=\!\!\!\!&
  \left[
    \begin{array}{c}
    -\sin\phi \sin\psi -\cos\phi \sin\theta \cos\psi \\
     \cos\phi \sin\psi -\sin\phi \sin\theta \cos\psi \\
    -\cos\theta \cos\psi 
    \end{array}
  \right]\!, \nonumber \\
  \negr v_3&\!\!\!\!=\!\!\!\!&
  \left[
    \begin{array}{c}
    -\cos\phi \cos\theta \\
    -\sin\phi \cos\theta \\
     \sin\theta 
    \end{array}
  \right]\!.
\end{eqnarray}

Finally, the axes of the intermediate joints are defined by the following unit vectors expressed in the base reference frame:
\begin{equation}
  \negr w_1\!=\!\!
  \left[
    \begin{array}{c}
    0 \\ \!\!\!-\sin\theta_1\!\! \\ ~\cos\theta_1\!\!
    \end{array}
  \right]\!, \,
  \negr w_2\!=\!\!
  \left[
    \begin{array}{c}
      ~\cos\theta_2\!\! \\ 0 \\ \!\!\!-\sin\theta_2\!\!
    \end{array}
  \right]\!, \,
  \negr w_3\!=\!\!
  \left[
    \begin{array}{c}
      \!\!\!-\sin\theta_3\!\! \\ ~\cos\theta_3\!\! \\ 0
    \end{array}
  \right]\!,\!\!
\end{equation}

\noindent
where $\theta_i$ is the \textit{active-joint variable} for leg $i$ (in this paper $i = 1, 2, 3$).

%%%%% SECTION 3 %%%%%%%%%%%%%%%%%%%%%%%%%%%%%%%%%%%%%%%%%%%%%%%%%%%%%%%%%%%%%%%%
\section{Inverse Kinematics and Working Modes}

Although very simple, the solutions to the inverse kinematics of the \textit{Agile Eye} will be presented for completeness \cite{Gosselin:1995}. For a given orientation of the mobile platform, each leg admits two solutions for $\theta_i$ in $(-\pi,\pi]$ obtained from
\begin{eqnarray}
  \tan\theta_1 \!\!\! &=& \!\!\! \frac{\cos\theta \sin\psi}{
                              \cos\phi \cos\psi +\sin\phi \sin\theta \sin\psi},\\
  \tan\theta_2 \!\!\! &=& \!\!\! \frac{\sin\phi \sin\psi +\cos\phi \sin\theta \cos\psi}
                              {\cos\theta \cos\psi},\\
  \tan\theta_3 \!\!\! &= &\!\!\! \tan\phi.
\end{eqnarray}

Thus, the inverse kinematic problem usually admits eight real solutions or working modes for any (non-singular) orientation of the mobile platform. It is very important to note that for both solutions for $\theta_i$, vector $\negr w_i$ is along the same axis but with opposite directions. When a leg is fully extended or fully folded, the corresponding equation from the above three ones does no hold true and the corresponding active-joint variable $\theta_i$ can be arbitrary.

%%%%% SECTION 4 %%%%%%%%%%%%%%%%%%%%%%%%%%%%%%%%%%%%%%%%%%%%%%%%%%%%%%%%%%%%%%%%
\section{Direct Kinematics and Assembly Modes}

The \textit{Agile Eye} was optimized to have maximum workspace and global dexterity \cite{Gosselin:1996}. Incidentally, such properties also yield great simplification in the direct kinematic problem because the eight solutions for a general 3-\textit{\underline{R}RR} parallel wrists degenerate to four trivial and four nontrivial ones. The direct kinematics of the {\em Agile Eye} was solved in \cite{Gosselin:1995} and will be reformulated and further analyzed here.

The following constraint equations are written:
\begin{subequations}
\begin{equation}
  \negr w^T_i \negr v_i = 0,
  \label{eq:contrainte_mgd}
\end{equation}
thus
\begin{eqnarray}
  && \sin\psi(\sin\theta_1 \sin\theta \sin\phi - \cos\theta \cos\theta_1) + \nonumber \\
  && \cos\psi \sin\theta_1 \cos\phi = 0, \\ 
  && \cos\psi(\cos\theta_2 \sin\theta \cos\phi - \cos\theta \sin\theta_2) + \nonumber\\
  && \sin\psi \cos\theta_2  \sin\phi = 0, \\
  && \sin(\theta_3-\phi) \cos\theta = 0.
\end{eqnarray}
\label{eq:mgd_123}
\end{subequations}
\vspace{-1em}

\noindent
From Eq.~(\ref{eq:mgd_123}d), the direct kinematic problem is found to admit two sets of solutions, defined by
\begin{subequations}
\begin{eqnarray}
&&\cos\theta = 0, \mbox{ and} \label{eq:set_sol_1}\\
  &&\sin(\theta_3-\phi) = 0. \label{eq:set_sol_2}
\end{eqnarray}
\end{subequations}

\noindent
In the next two subsections, these two equations will be solved.

%%%%% SUBSECTION 4.1 -----------------------------------------------------------
\subsection{First Set of Solutions --- Trivial Solutions}

Equation~(\ref{eq:set_sol_1}) gives two solutions for the angle $\theta$,
\begin{equation}
  \theta = \pi/2 \quad {\rm and} \quad \theta = -\pi/2,
\end{equation}

\noindent
which both correspond to a representation singularity in the $XYZ$ Euler angles. From the first solution and after simplification of Eq.~(\ref{eq:mgd_123}b), the following condition is found for arbitrary active-joint variables,
\begin{equation}
  \cos(\phi-\psi)=0,
\end{equation}
and from the second solution,
\begin{equation}
  \cos(\phi+\psi)=0.
\end{equation}

In both cases, each of these two equations lead to two solutions. Thus, because of the representation singularity, only four rotation matrices describe the corresponding orientations of the mobile platform:
\begin{eqnarray*}
&&\!\!\!\!\!\!\!\!\!\!\!\!\!\!\!\negr R_{TO1} = 
 \left[
   \begin{array}{ccc}
   0        & \!\!\!-1 & 0 \\
   0        &        0 & 1 \\
   \!\!\!-1 &        0 & 0 
   \end{array}
 \right] {\rm ,~}
\negr R_{TO2} = 
 \left[
   \begin{array}{ccc}
       0 &  1 &       0 \\
       0 &  0 &\!\!\!-1 \\
\!\!\!-1 &  0 &       0 
   \end{array}
 \right] {\rm ,~}\\
&&\!\!\!\!\!\!\!\!\!\!\!\!\!\!\!\negr R_{TO3} = 
 \left[
   \begin{array}{ccc}
   0  & \!\!\!-1 &       0 \\
   0  &        0 &\!\!\!-1 \\
   1  &        0 &       0 
   \end{array}
 \right] {\rm ,~}
\negr R_{TO4} = 
 \left[
   \begin{array}{ccc}
   0  &  1 & 0 \\
   0  &  0 & 1 \\
    1 &  0 & 0 
   \end{array}
 \right].
\label{eq:orientation_trivial}
\end{eqnarray*}

%%%%% FIGURE 2 %%%%%%%%---------------------------------------------------------
\begin{figure}[t]
  \begin{center}
    \subfigure[$\theta = \pi/2$, $\phi-\psi = \pi/2$]{
     \scalebox{0.21}{\includegraphics{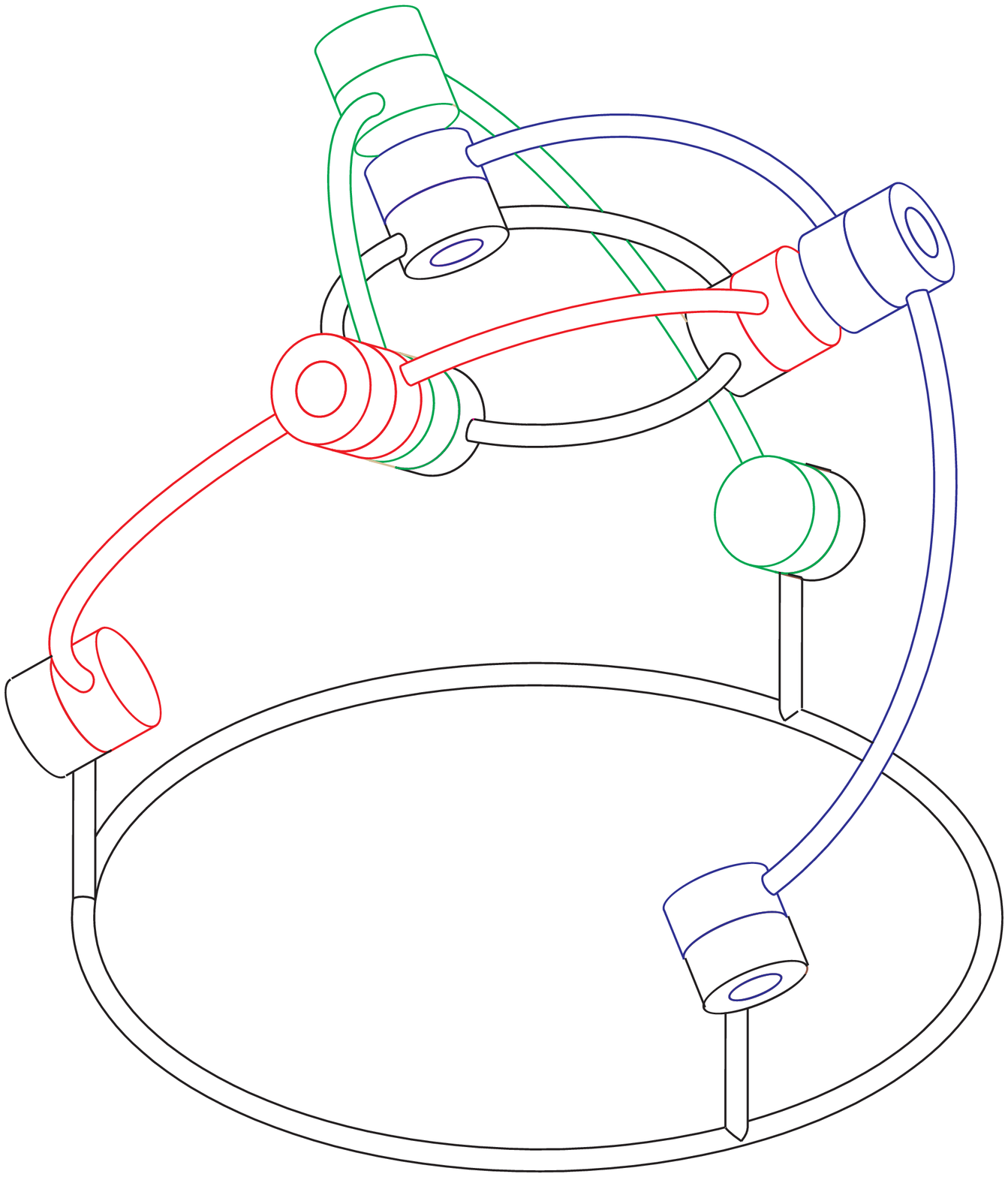}}} \hspace{5mm}
    \subfigure[$\theta = \pi/2$, $\phi-\psi = -\pi/2$]{
     \scalebox{0.21}{\includegraphics{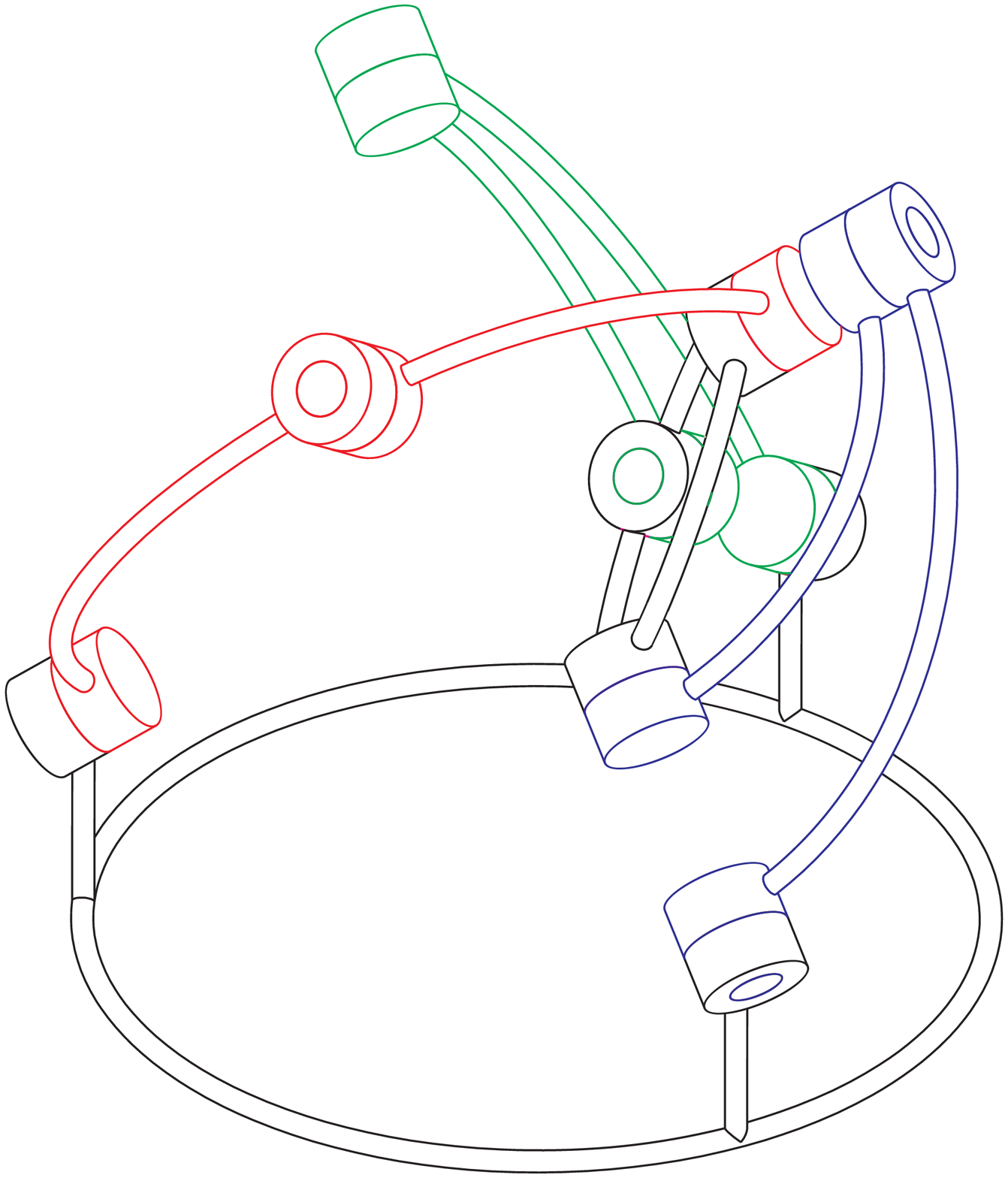}}} \\
    \subfigure[$\theta = -\pi/2$, $\phi+\psi = \pi/2$]{
     \scalebox{0.21}{\includegraphics{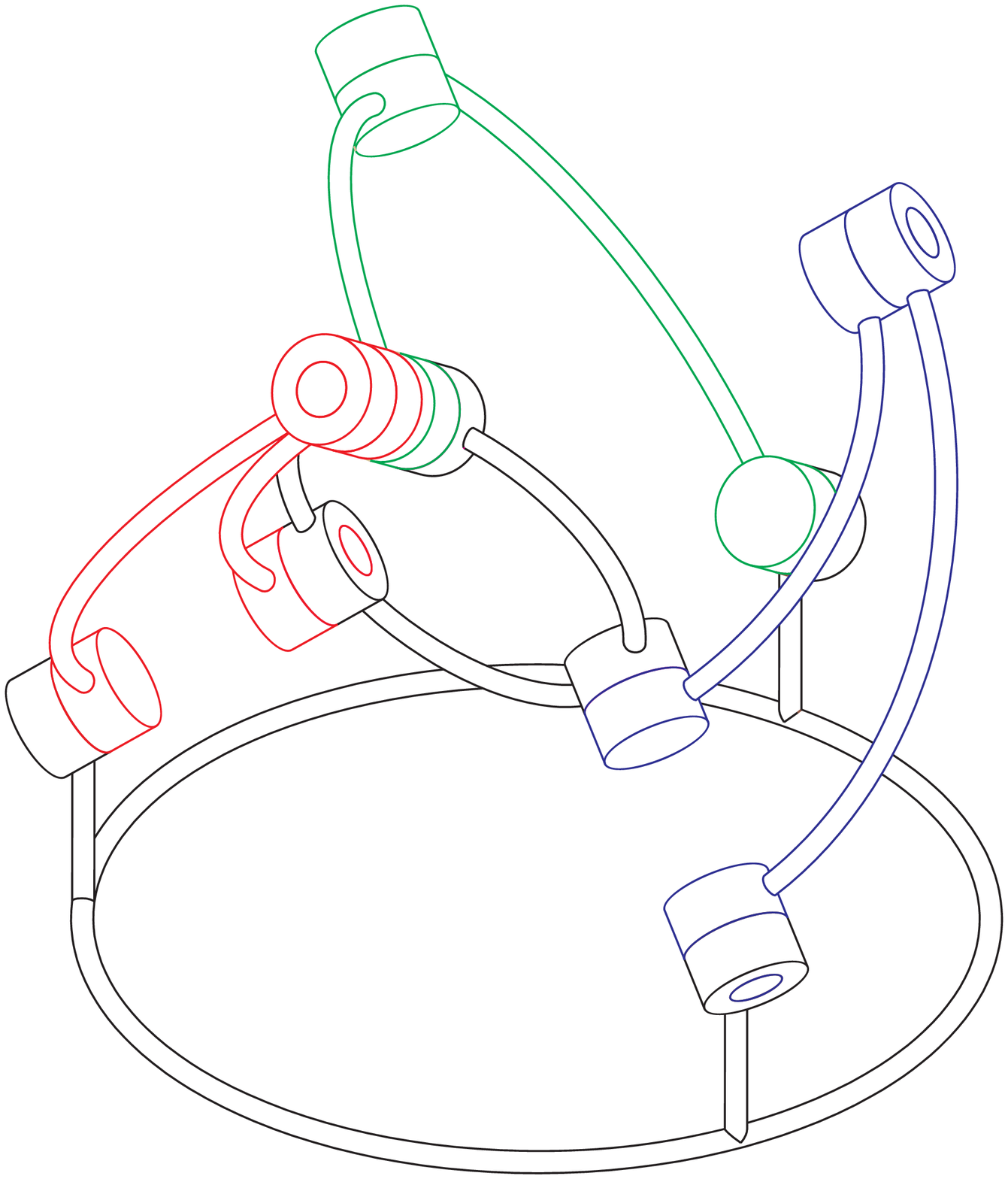}}} \hspace{5mm}
    \subfigure[$\theta = -\pi/2$, $\phi+\psi = -\pi/2$]{
     \scalebox{0.21}{\includegraphics{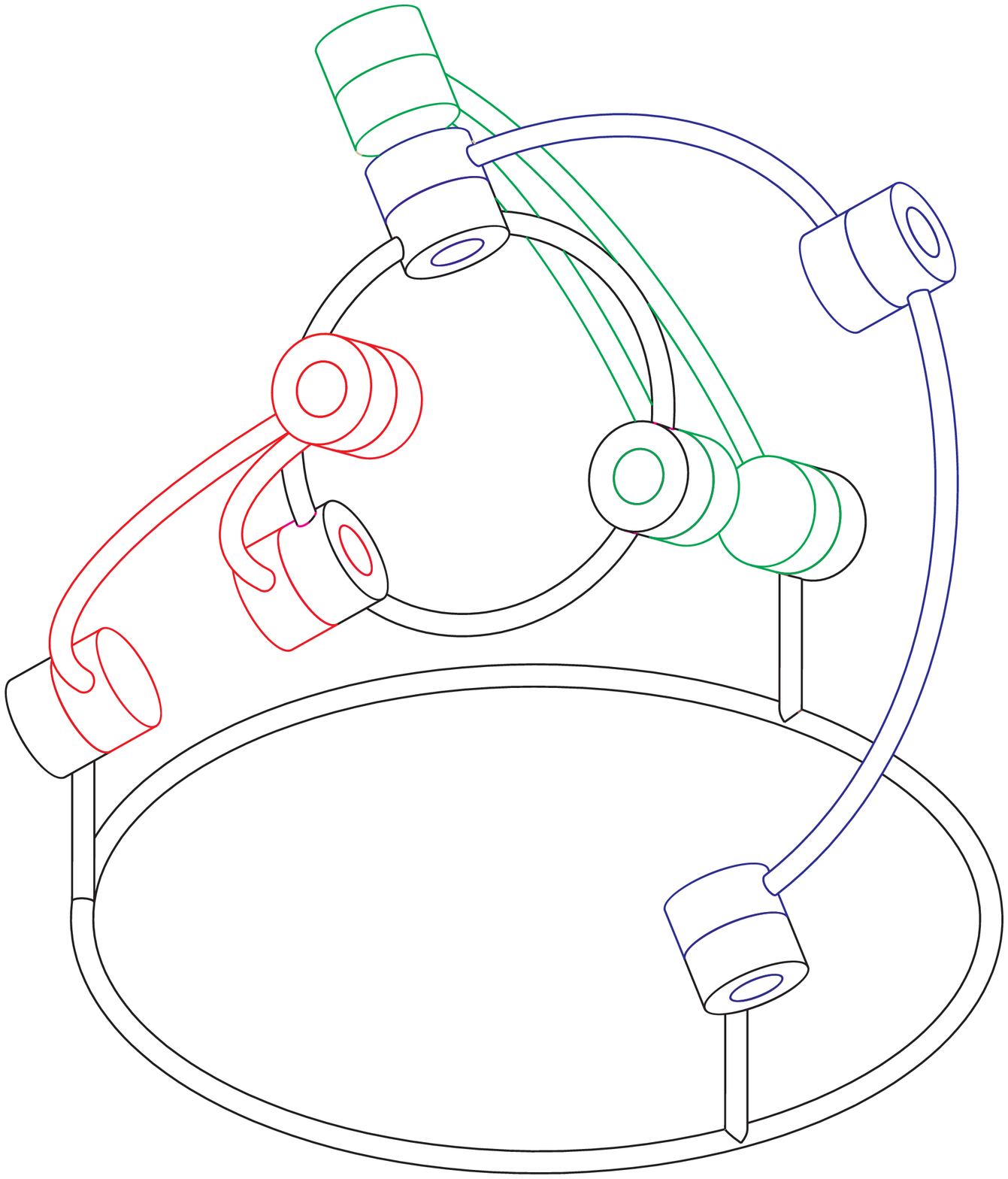}}} 
    \caption{The four trivial solutions to the direct kinematic problem valid
             for any set of active-joint variables but shown with $\theta_1 = 0$,
             $\theta_2 = 0$ and $\theta_3 = 0$.}
    \protect\label{fig:Trivial}
  \end{center}
  \vspace*{-5mm}
\end{figure}
%%%%%%%%%%%%%%%%%%%%%%%---------------------------------------------------------

Figure~\ref{fig:Trivial} depicts the four trivial solutions (orientations) to the direct kinematic problem. It may be seen geometrically, as well as from Eqs.~(\ref{eq:mgd_123}b-d), that when the platform is at one of these four orientations, all three legs are at a singularity (fully extended or folded) and can freely rotate about their base joint axes. Thus, these four orientations are trivial solutions to the direct kinematic problem of the \textit{Agile Eye} and exist for any set of active-joint variables. These four families of configurations will be discussed further in the section on singularity analysis.

%%%%% SUBSECTION 4.2 -----------------------------------------------------------
\subsection{Second Set of Solutions --- Nontrivial Solutions}

Equation~(\ref{eq:set_sol_2}) gives two solutions for the angle $\phi$:
\begin{equation}
  \phi=\theta_3 \quad {\rm and} \quad \phi=\theta_3\pm\pi.
\end{equation}

\noindent
However, in the \textit{ZYX} Euler-angle convention, the triplets $\{\phi, \theta, \psi\}$ and $\{\phi \pm \pi, -\theta\pm \pi, \psi \pm \pi\}$  both correspond to the same orientation. This means that the above two solutions will lead to the same orientation of the mobile platform. Hence, only the first solution will be used in this paper. 

Substituting $\phi=\theta_3$ in Eqs.~(\ref{eq:mgd_123}b-c), a new system of two equations is obtained:
\begin{subequations}
\begin{eqnarray}
  p_1 \cos \psi + p_2 \sin \psi&=&0, \label{eq:p1_p2}\\
  p_3 \cos \psi + p_4 \sin \psi&=&0,
\end{eqnarray}
where
\vspace*{-2mm}
\begin{eqnarray}
 p_1 \!&=&\! \sin\theta_1 \cos\theta_3, \label{equations:p1p2p3p4}\\
 p_2 \!&=&\! \sin\theta_1 \sin\theta \sin\theta_3 - \cos\theta \cos\theta_1, \\
 p_3 \!&=&\! \cos\theta_2 \sin\theta \cos\theta_3 - \cos\theta \sin\theta_2, \\
 p_4 \!&=&\! \cos\theta_2 \sin\theta_3.
\end{eqnarray}
\end{subequations}

\noindent
Since the terms $\cos\psi$ and $\sin\psi$ cannot vanish simultaneously, Eqs.~(\ref{eq:p1_p2}-b) lead to
\vspace*{-2mm}
\begin{equation}
  p_1p_4-p_2p_3=0.
\end{equation}

\noindent
Hence, substituting Eqs.~(\ref{equations:p1p2p3p4}-f) into the above equation gives
\begin{subequations}
\begin{equation}
  \cos\theta ( q_1 \cos\theta + q_2 \sin\theta)=0, \label{eq:sol_theta}
\end{equation}
where
\vspace*{-1mm}
\begin{eqnarray}
\!\!q_1\!\!&=&\!\!
      \sin\theta_1 \cos\theta_2 \cos\theta_3 \sin\theta_3 - \cos\theta_1 \sin\theta_2,\\
\!\!q_2\!\!&=&\!\!
      \sin\theta_1 \sin\theta_2 \sin\theta_3 + \cos\theta_1 \cos\theta_2 \cos\theta_3.
\end{eqnarray}
\end{subequations}

\noindent
Equation~(\ref{eq:sol_theta}) leads to two possibilities:
\begin{subequations}
\begin{eqnarray}
  \cos\theta= 0 \quad {\rm~and} \label{eq:first_sol_theta}\\
  q_1 \cos\theta + q_2 \sin\theta=0. \label{eq:second_sol_theta}
\end{eqnarray}
\end{subequations}

\noindent
Equation~(\ref{eq:first_sol_theta}) is the same as Eq.~(\ref{eq:set_sol_1}) and will therefore be discarded. Equation~(\ref{eq:second_sol_theta}) gives two solutions in $(-\pi,\pi]$:
\begin{equation}
  \theta = \tan^{-1}(-q_1/q_2) + k\pi {\rm~with~} k=0,1.
\end{equation}

Equations~(\ref{eq:p1_p2}-b) can now be used to find $\psi$:
\begin{eqnarray*}
  \psi \!\!&=&\!\! \tan^{-1}(-p_1/p_2)+ k\pi {\rm~with~} k=0,1, {\rm~or}\\
  \psi \!\!&=&\!\! \tan^{-1}(-p_3/p_4)+ k\pi {\rm~with~} k=0,1,
\end{eqnarray*}
which gives two values for $\psi$ in $(-\pi,\pi]$ for each $\theta$.

\begin{figure}[t]
  \begin{center}
    \subfigure[$\phi = 0.100$, $\theta = -0.672$, $\psi = -0.383$]
    {\scalebox{0.2}{\includegraphics{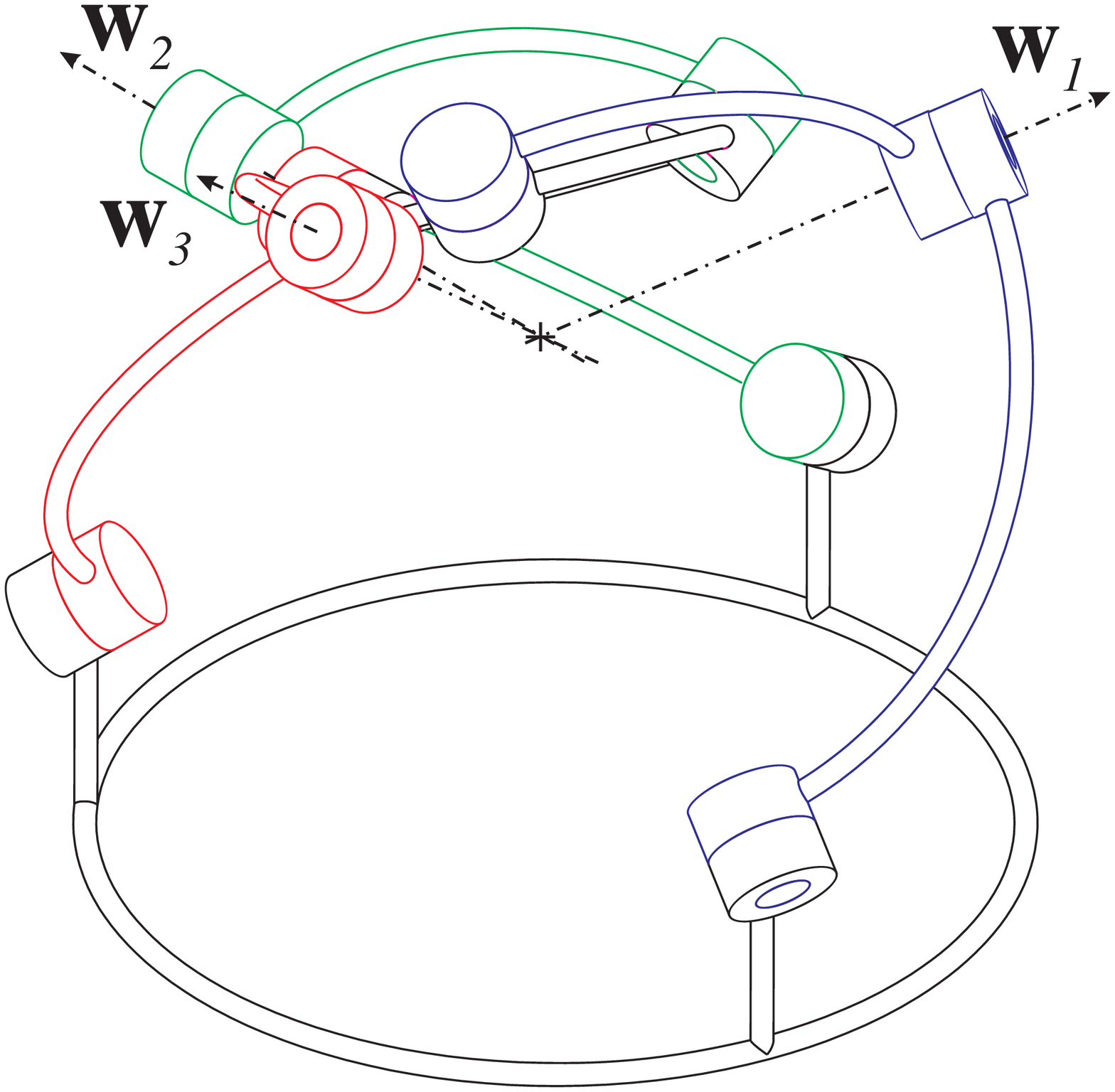}}}\hspace{5mm}
    \subfigure[$\phi = 0.100$, $\theta = -0.672$, $\psi = 2.759$]
    {\scalebox{0.2}{\includegraphics{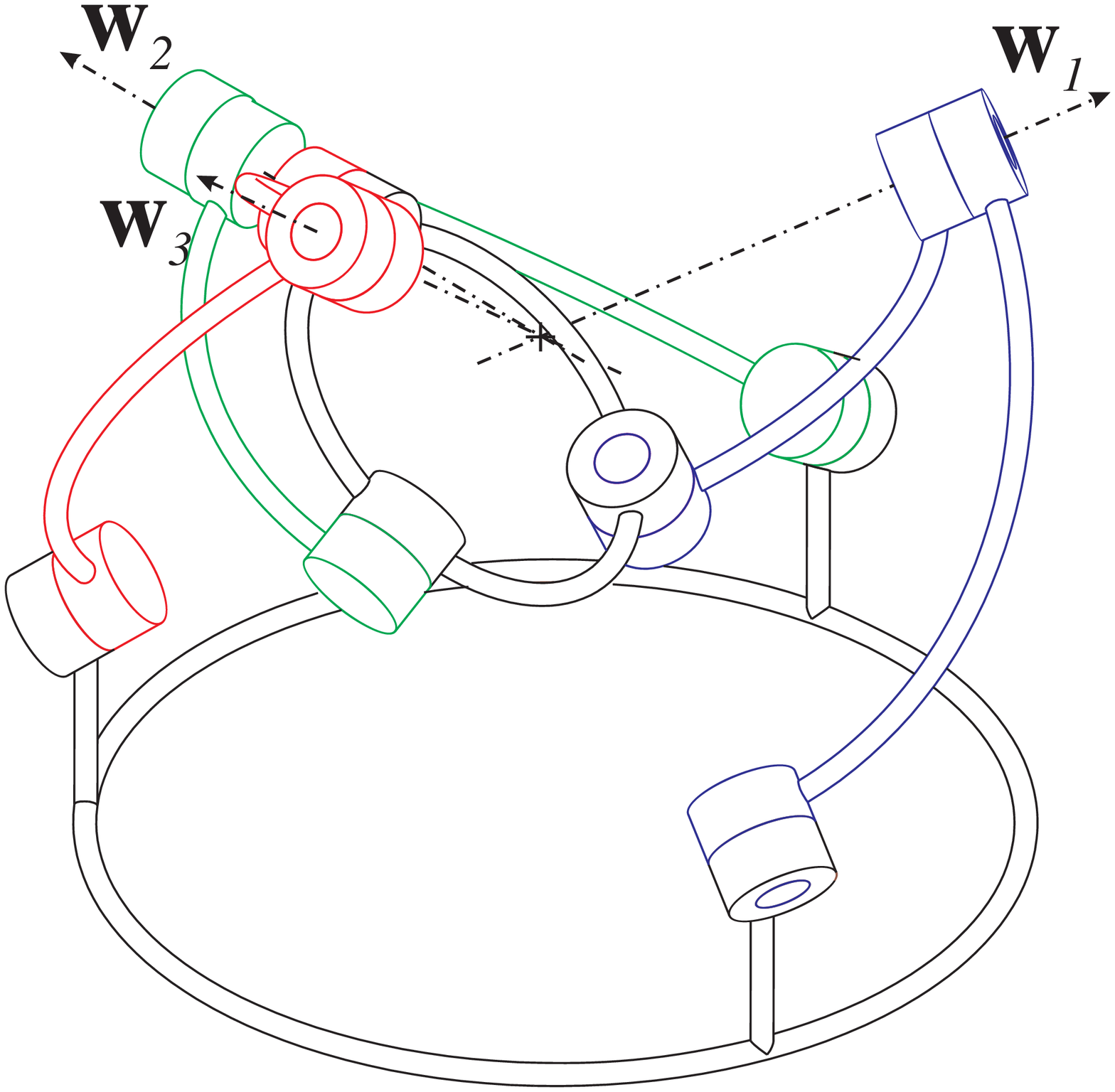}}}\\
    \subfigure[$\phi = 0.100$, $\theta = 2.470$, \mbox{$\psi = 0.383$}]
    {\scalebox{0.2}{\includegraphics{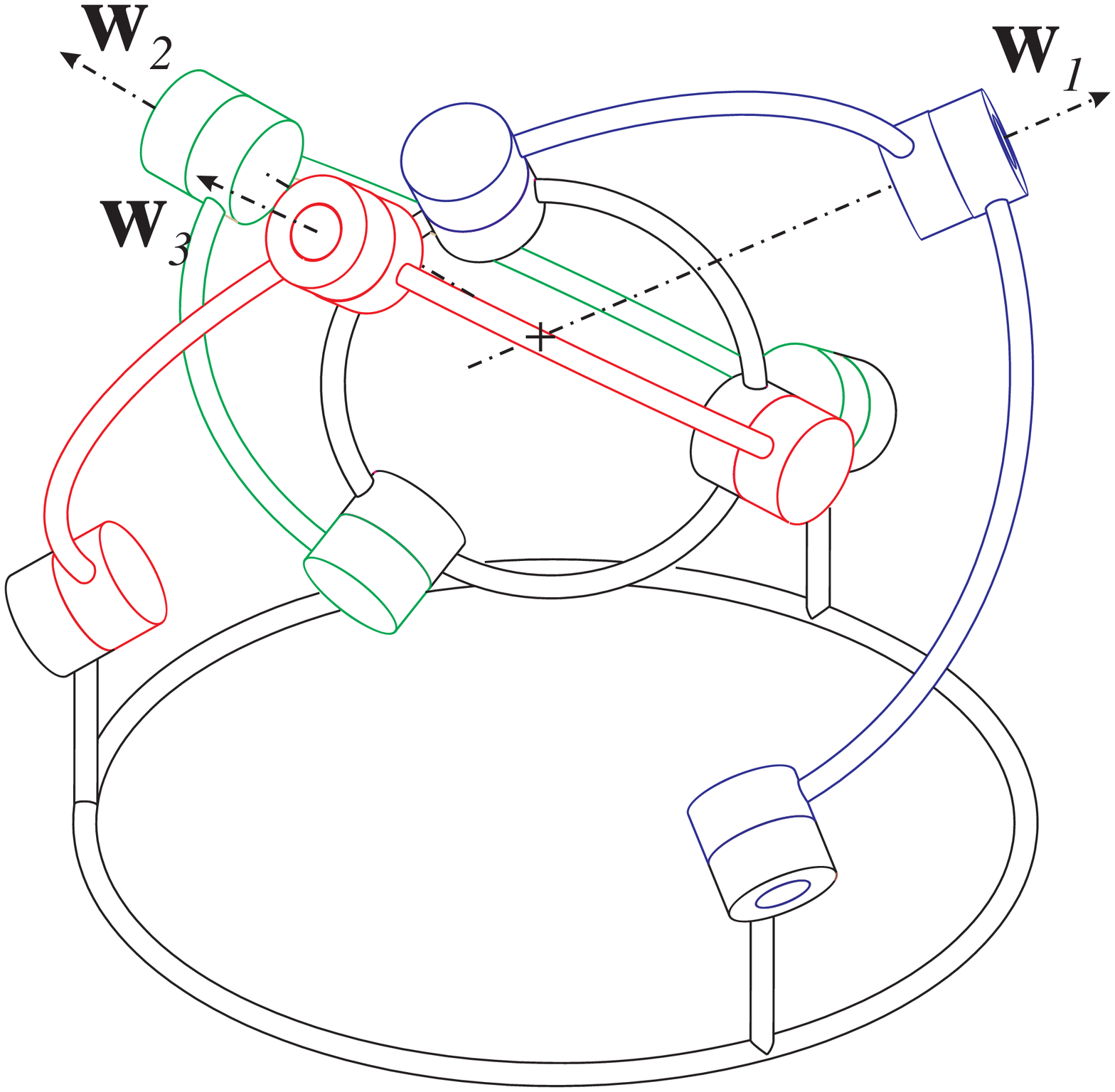}}}\hspace{5mm}
    \subfigure[$\phi = 0.100$, $\theta = 2.470 $, \mbox{$\psi = 3.525$}]
    {\scalebox{0.2}{\includegraphics{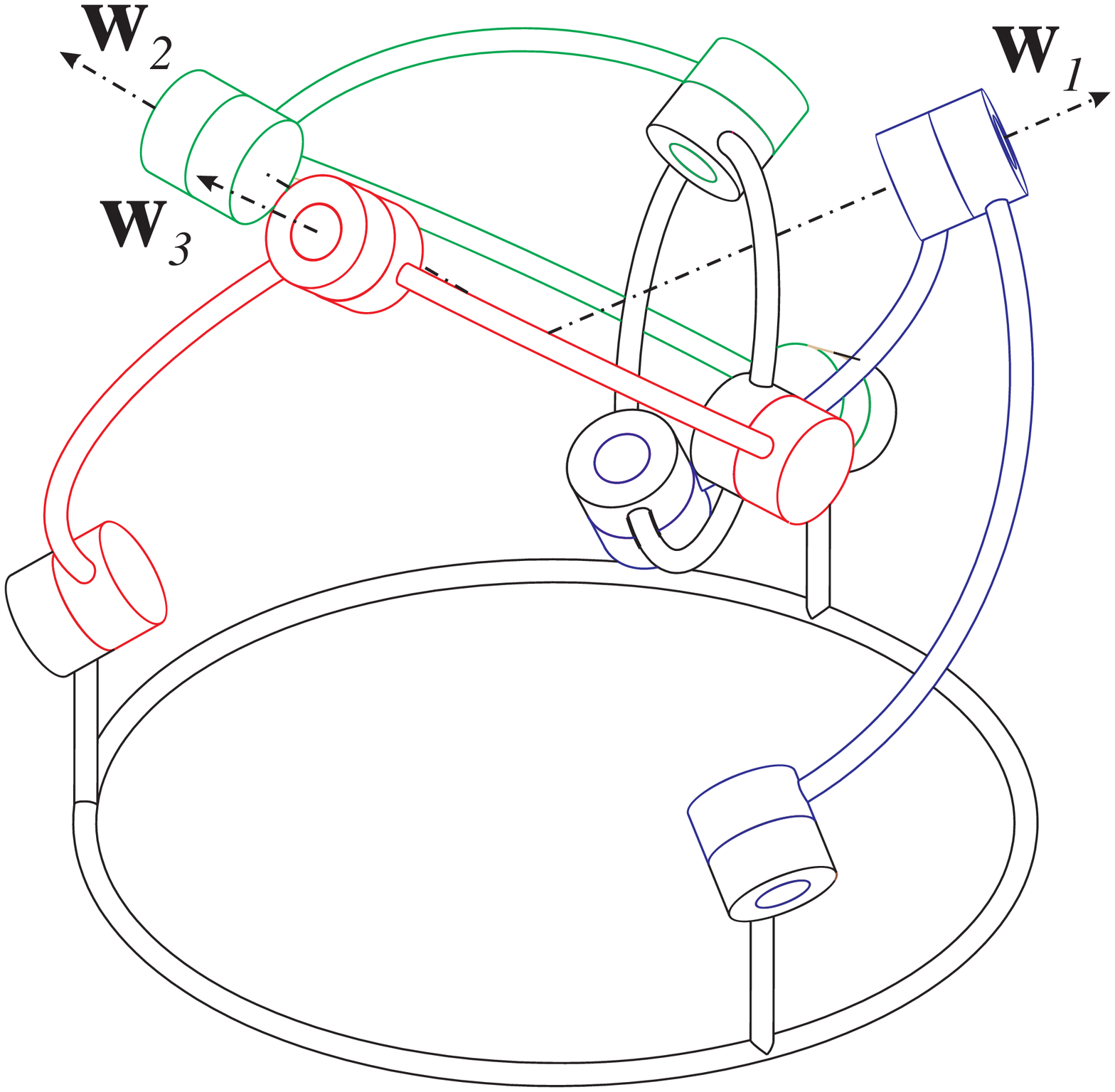}}}
    \caption{The four nontrivial solutions to the direct kinematic problem of the
             \textit{Agile Eye} for $\theta_1=-0.3, \theta_2=-0.7$ and $\theta_3=0.1$.}
    \label{fig:Solution_DKP}
  \end{center}
\end{figure}

Table~\ref{table_solution_DKP} summarizes the solutions to the direct kinematic problem, where a number to each solution is arbitrarily assigned. An example of four nontrivial solutions to the direct kinematic problem of the \textit{Agile Eye} is given in Fig.~\ref{fig:Solution_DKP} for the active-joint variables $\theta_1=-0.3, \theta_2=-0.7$ and $\theta_3=0.1$.

Later, a method will be presented to identify these nontrivial solutions based on the working mode. For now, note by solely observing Fig.~\ref{fig:Solution_DKP} that these four nontrivial solutions are obtained from each other by rotating the mobile platform about a platform joint axis at 180 degrees (in fact, the same is true for the four trivial solutions). Thus, if for a given set of active-joint variables, there is one nontrivial solution, then there are (at least) three other nontrivial solutions. The important question whether there are always four nontrivial solutions was not answered in \cite{Gosselin:2002} and will be given special attention now.

%%%%% FIGURE 3 %%%%%%%%---------------------------------------------------------
%%%%%%%%%%%%%%%%%%%%%%%---------------------------------------------------------

%%%%% TABLE 1 %%%%%%%%----------------------------------------------------------
\begin{table}[t]
   \center
   \renewcommand{\arraystretch}{1.3}
   \begin{tabular}{|c||c|c|c|} \hline
      Solution 1 & $\phi$ & $\theta$     & $\psi$      \\ \hline
      Solution 2 & $\phi$ & $\theta$     & $\psi+\pi$  \\ \hline
      Solution 3 & $\phi$ & $\theta+\pi$ & $-\psi$     \\ \hline
      Solution 4 & $\phi$ & $\theta+\pi$ & $-\psi+\pi$ \\ \hline
   \end{tabular}
   \renewcommand{\arraystretch}{1}
   \caption{The four nontrivial solutions to the direct kinematic problem}
   \label{table_solution_DKP}
\end{table}
%%%%%%%%%%%%%%%%%%%%%%%---------------------------------------------------------

%%%%% SUBSECTION 4.3 -----------------------------------------------------------
\subsection{Degenerate Cases}

The second set of solutions will become the same as the first set of solutions when
Eq.~(\ref{eq:second_sol_theta}) degenerates and $q_2 = q_1 = 0$. It can be shown that $q_2 = q_1 = 0$ if and only if $\sin\theta_2=0$ and $\cos\theta_3=0$, or $\sin\theta_3=0$ and $\cos\theta_1=0$. In that case, $\theta$ can be anything, which is one of the self-motions of the mobile platform. If $q_2 = 0$ but $q_1 \ne 0$, Eq.~(\ref{eq:second_sol_theta}) becomes identical with Eq.~(\ref{eq:first_sol_theta}), meaning that $\theta=\pm\pi/2$. Substituting $\theta=\pm\pi/2$ into Eqs.~(\ref{eq:p1_p2}-b) yields
\begin{equation}
  \sin\theta_1\cos(\phi\mp\psi) = 0 \mbox{  and   } \cos\theta_2\cos(\phi\mp\psi) = 0.
\end{equation}

\noindent
If $\sin\theta_1 = 0$ and $\cos\theta_2 = 0$, ($\phi\mp\psi$) is arbitrary, meaning that the platform can undergo a self motion. If, however, these two conditions are not satisfied, then $\cos(\phi\mp\psi) = 0$, which means that the only direct kinematic solutions are the trivial ones.

Thus, in summary, the \textit{Agile Eye} will have only the four trivial solutions to its direct kinematic problem if and only if
\begin{equation}
 q_2 = \sin\theta_1\sin\theta_2\sin\theta_3 +
 \cos\theta_1\cos\theta_2\cos\theta_3 = 0, \label{eq:q2zero}
\end{equation}
but neither of the following three pairs of conditions is true:
\begin{subequations}
\begin{eqnarray}
  &&\sin\theta_2 = 0 \mbox{  and   } \cos\theta_3 = 0, \mbox{  or}\label{eq:cond1}\\
  &&\sin\theta_3 = 0 \mbox{  and   } \cos\theta_1 = 0, \mbox{  or}\label{eq:cond2}\\
  &&\sin\theta_1 = 0 \mbox{  and   } \cos\theta_2 = 0.\label{eq:cond3}
\end{eqnarray}
\end{subequations}

Taking the first pair yields $q_2 = 0$ and $q_1 = 0$, meaning that $\theta$ can take any value. This means that there is a self motion --- even if all actuators are fixed, the platform can freely move. Substituting $\theta_2=0$ or $\theta_2=\pi$, and $\theta_3=\pm\pi/2$ into Eqs.~(\ref{eq:p1_p2}-b), yields $\sin\psi = 0$. In addition, $\cos\phi=\cos\theta_3=0$. Similarly, for the second pair of conditions, it can be proved that $\theta$ can have any value, while $\cos\psi=0$ and $\sin\phi=0$. Finally, taking the third pair of conditions, $q_2 = 0$ but $q_1 \ne 0$, Eq.~\ref{eq:sol_theta} implies that $\cos\theta = 0$. If this pair of conditions is substituted into Eqs.~(\ref{eq:p1_p2}-b), it is reached to the conclusion that $\sin(\phi-\psi)$ can be anything in the case of $\theta = \pi/2$ or that $\sin(\phi+\psi)$ can be anything in the case of $\theta = -\pi/2$.

Thus, it can be easily shown that Eqs.~(\ref{eq:cond1}--c) correspond to six self motions represented by the following rotation matrices (where angles are arbitrary):
\begin{subequations}
\begin{eqnarray}
\renewcommand{\arraystretch}{1.5}
  \negr R_{SM1a} \!\!\!&=&\!\!\! \left[\begin{array}{ccc}
                            0 & -1 &          0 \\
                   \cos\theta &  0 & \sin\theta \\
                  -\sin\theta &  0 & \cos\theta
             \end{array}\right]\!\!,\label{eq:selfmotion1}\\
  \negr R_{SM1b} \!\!\!&=&\!\!\! \left[\begin{array}{ccc}
                            0 &  1 &          0 \\
                   \cos\theta &  0 & -\sin\theta \\
                  -\sin\theta &  0 & -\cos\theta
             \end{array}\right]\!\!,\label{eq:selfmotion2}\\
  \negr R_{SM2a} \!\!\!&=&\!\!\! \left[\begin{array}{ccc}
                   \cos\theta & \sin\theta &  0 \\
                            0 &          0 & -1 \\
                  -\sin\theta & \cos\theta &  0
             \end{array}\right]\!\!,\label{eq:selfmotion3}\\
  \negr R_{SM2b} \!\!\!&=&\!\!\! \left[\begin{array}{ccc}
                   \cos\theta & -\sin\theta & 0 \\
                            0 &           0 & 1 \\
                  -\sin\theta & -\cos\theta & 0
             \end{array}\right]\!\!,\label{eq:selfmotion4}\\
  \negr R_{SM3a} \!\!\!&=&\!\!\! \left[\begin{array}{ccc}
                   0 & -\sin(\phi-\psi) &  \cos(\phi-\psi) \\
                   0 &  \cos(\phi-\psi) &  \sin(\phi-\psi) \\
                  -1 &                0 &               0  
             \end{array}\right]\!\!,\label{eq:selfmotion5}\\
  \negr R_{SM3a} \!\!\!&=&\!\!\! \left[\begin{array}{ccc}
                  0 & -\sin(\phi+\psi) & -\cos(\phi+\psi) \\
                  0 &  \cos(\phi+\psi) & -\sin(\phi+\psi) \\
                  1 &                0 &               0  
              \end{array}\right]\!\!.\label{eq:selfmotion6}
\renewcommand{\arraystretch}{1}
\end{eqnarray}
\end{subequations}

\begin{figure}[t]
  \begin{center}
    \subfigure[]{\scalebox{0.21}{\includegraphics{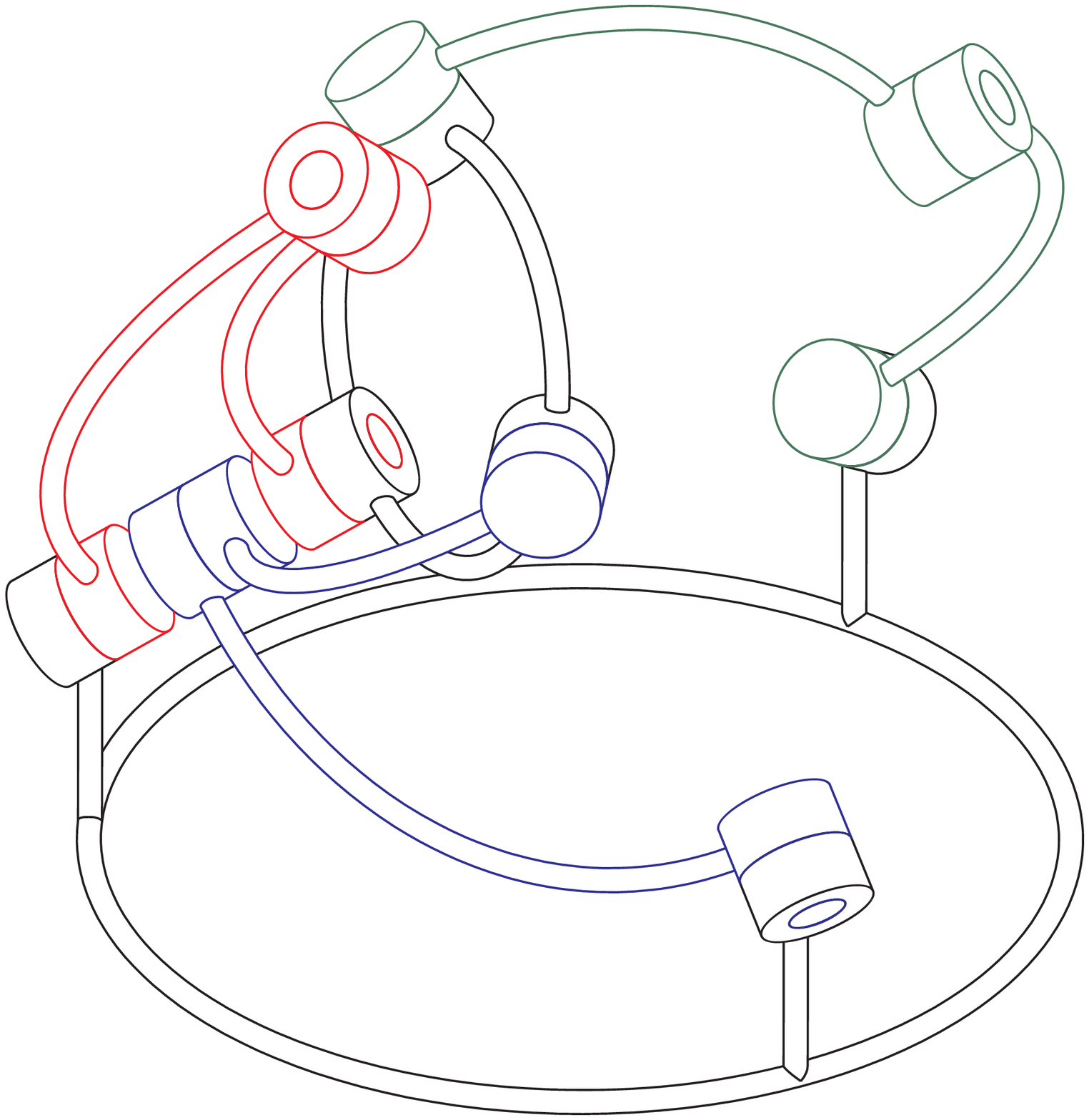}}} \hspace{5mm}
    \subfigure[]{\scalebox{0.21}{\includegraphics{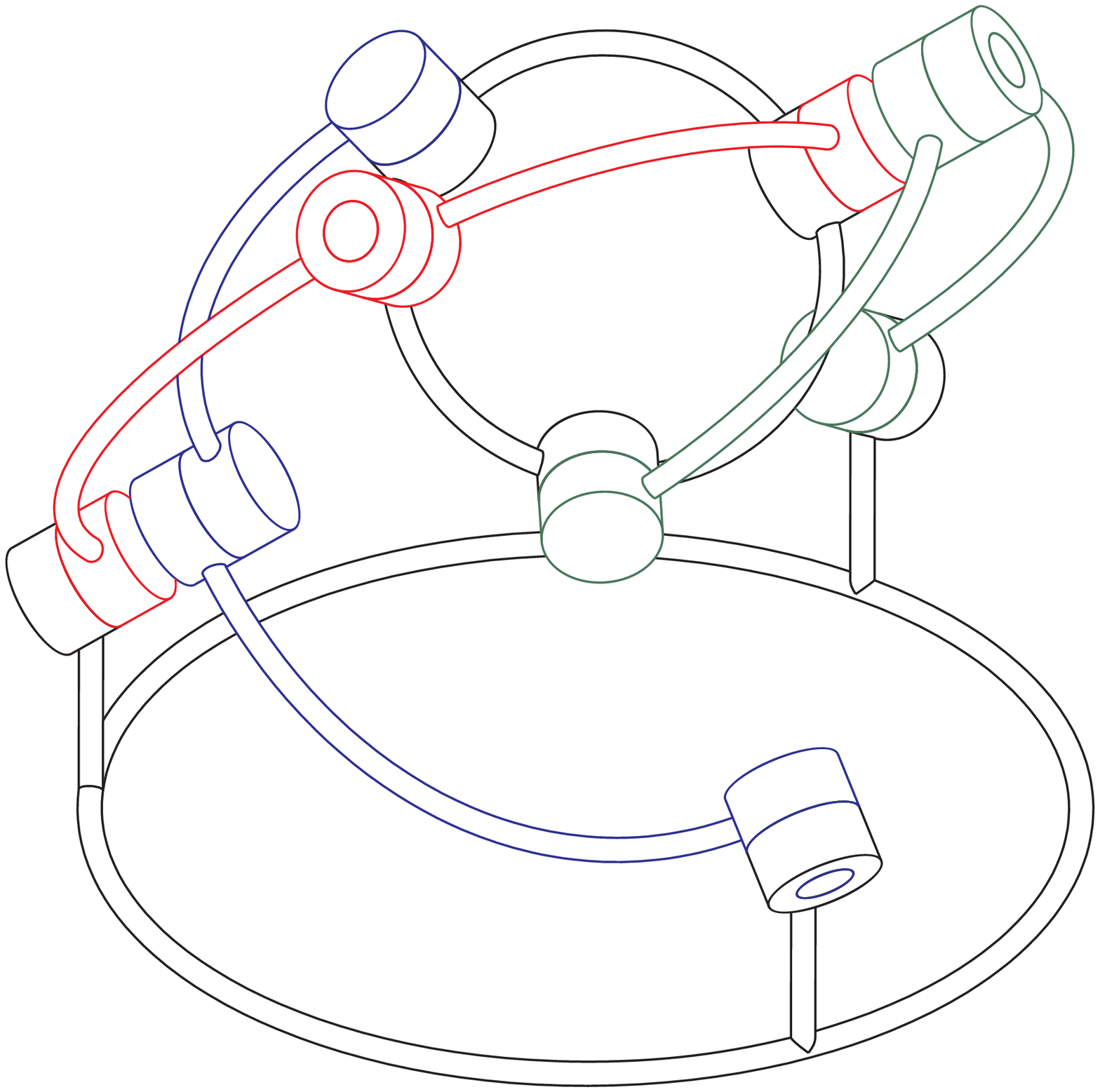}}}
    \caption{The two self motions of the mobile platform where leg 3 is singular.}
    \label{fig:SelfMotions}
  \end{center}
\end{figure}

Now, note that each of the pairs of conditions imposes a constraint on two of the active-joint variables, while the third one can take any value, without influencing the orientation of the mobile platform. This means, that in the above self-motions, there is a leg in singularity. In fact, the self-motion of the platform is about the axis of the base joint of the leg in singularity. The above self motions are divided into pairs, where in each pair, one of the motions correspond to a fully extended leg (the ones with the b index), while the other to a fully folded one. Figure~\ref{fig:SelfMotions} shows the two self motions (SM3a and SM3b) for which leg 3 is singular, corresponding to the pair of conditions of Eq.~(\ref{eq:cond3}). In this figure, legs 1 and 2 are each shown in one of the possible two configurations per leg.

%%%%% FIGURE 4 %%%%%%%%---------------------------------------------------------
%%%%%%%%%%%%%%%%%%%%%%%---------------------------------------------------------

As will be seen in the next section, all singular configurations were found in this section by purely studying the degeneracies of the direct kinematics of the \textit{Agile Eye}.

%%%%% SECTION 5 -----------------------------------------------------------
\section{Singularity Analysis}

The relationship between the active-joint rates, $\dot {\gbf \theta}$, and the angular velocity of the mobile platform, $\gbf \omega$,  can be written as:
\begin{equation}
  \negr A \gbf \omega + \negr B \dot{\gbf \theta} = 0
  \label{eq:A_B}
\end{equation}
where $\negr A$ and $\negr B$ are Jacobian matrices and can be written as
\begin{subequations}
\begin{equation}
\negr A= 
\left[
  \begin{array}{c}
     (\negr w_1 \times \negr v_1)^T \\
     (\negr w_2 \times \negr v_2)^T \\
     (\negr w_3 \times \negr v_3)^T 
  \end{array}
\right] =
\left[
  \begin{array}{c}
      \gbf \alpha_1^T \\
      \gbf \alpha_2^T \\
      \gbf \alpha_3^T 
  \end{array}
\right],
\label{eq:matrix_A}
\end{equation}
\begin{equation}
\negr B= \!
\left[\!\!
  \begin{array}{ccc}
   (\negr w_1 \times \negr v_1)^T \negr u_1
   \!\!\!\! & 
   0 & 
   0\\
   0 & 
   \!\!\!
   (\negr w_2 \times \negr v_2)^T \negr u_2
   \!\!\!\! &
   0\\
   0 & 
   0 & 
   \!\!\!\!
   (\negr w_3 \times \negr v_3)^T \negr u_3
  \end{array}
\!\!\right]\!.\!
\label{eq:matrix_B}
\end{equation}
\end{subequations}

Type~2 singularities are characterized by studying matrix $\negr A$ and occur whenever the three vectors $\gbf \alpha_i$ are coplanar or collinear. For the {\em Agile Eye}, these vectors cannot be collinear. Thus, when these three vectors are coplanar, the platform can rotate (infinitesimally or finitely) about the axis passing through the center $O$ and normal to the plane of the vectors.

Substituting the nontrivial solution set for the direct kinematics of the \emph{Agile Eye} into the determinant of $\negr A$ of Eq.~(\ref{eq:matrix_A}) and simplifying, it is obtained that the expression in the active-joint space for Type~2 singularities is:
\begin{equation}
  \det(\negr A) = \sin\theta_1\sin\theta_2\sin\theta_3+\cos\theta_1\cos\theta_2\cos\theta_3 = 0.\label{eq:Azero}
\end{equation}

Substituting the trivial solution set yields the same expression but with a negative sign. Note that the determinant of $\negr A$ is the same for all four assembly modes. This is not surprising since from the geometric interpretation of the four assembly modes, it can be seen that between any two assembly modes, two pairs of vectors $\gbf \alpha_i$ have opposite directions and the third pair is the same.

Type~1 singularities are characterized by studying matrix $\negr B$ and occur whenever a leg is fully extended or folded. For a general 3-\textit{\underline{R}RR} spherical parallel mechanism with legs of other than 90 degrees, Type~1 singularities are two-dimensional entities. In other words, for such a general parallel wrist, when a single leg is singular, the platform remains with two degrees of freedom. In the \textit{Agile Eye}, however, when a leg is singular, the axes of the base and platform joints coincide and the mobile platform has a single degree of freedom, whereas the leg can freely rotate without affecting the orientation of the mobile platform. Therefore, Type~1 singularities of the \textit{Agile Eye} are only six curves in the orientation space.

Substituting the nontrivial solution set for the direct kinematics of the \emph{Agile Eye} into $\negr B$ of Eq.~(\ref{eq:matrix_B}) and simplifying yields the following three expressions in the active-joint space, corresponding to Type~1 singularities occurring in leg 1, 2, and 3, respectively:
{\jot=10pt \begin{eqnarray}
  \!\!\!\!\!\!\!B_{11} = \pm \frac{\sin\theta_1\sin\theta_2\sin\theta_3+
        \cos\theta_1\cos\theta_2\cos\theta_3}{
  \sqrt{1-\cos^2\theta_3\sin^2\theta_1}\sqrt{1-\cos^2\theta_1\sin^2\theta_2}}
  \!\!\!\!\!&=&\!\!\!\!\! 0,\label{eq:B11zero}\\
  \!\!\!\!\!\!\!B_{22} = \pm \frac{\sin\theta_1\sin\theta_2\sin\theta_3+
        \cos\theta_1\cos\theta_2\cos\theta_3}{
  \sqrt{1-\cos^2\theta_2\sin^2\theta_3}\sqrt{1-\cos^2\theta_1\sin^2\theta_2}}
  \!\!\!\!\!&=&\!\!\!\!\! 0,\label{eq:B22zero}\\
  \!\!\!\!\!\!\!B_{33} = \pm \frac{\sin\theta_1\sin\theta_2\sin\theta_3+
        \cos\theta_1\cos\theta_2\cos\theta_3}{
  \sqrt{1-\cos^2\theta_2\sin^2\theta_3}\sqrt{1-\cos^2\theta_3\sin^2\theta_1}}
  \!\!\!\!\!&=&\!\!\!\!\! 0,\label{eq:B33zero}
\end{eqnarray}}

\noindent
where $B_{ii}$ is the $i$-th diagonal element of $\negr B$, and the plus-minus sign depends on which of the four nontrivial direct solutions is used, i.e., on the assembly mode.

Substituting the trivial solution set for the direct kinematics of the \emph{Agile Eye} into $\negr B$ and simplifying yields as expected:
\begin{equation}
  B_{11} = B_{22} = B_{33} = 0.\label{eq:Biizero}
\end{equation}
Indeed, at the trivial orientations all three legs are singular.

From Eqs.~(\ref{eq:B11zero}--\ref{eq:B33zero}), it follows that if a configuration corresponds to a Type~2 singularity, then it should inevitably correspond to a Type~1 singularity too. However, the opposite is not necessarily true. In other words, if there is a Type~1 singularity, the \emph{Agile Eye} is not necessarily at a Type~2 singularity too. Indeed, investigating the four orientations shown in Fig.~\ref{fig:Trivial}, it can be seen that the legs can be orientated in such a way that the vectors normal to the last two joint axes in each leg are not coplanar. Such a configuration is called a lockup configuration, since the mobile platform is completely restrained even if the actuators are removed. For these four orientations, Eqs.~(\ref{eq:B11zero}--\ref{eq:B33zero}) are replaced by Eq.~(\ref{eq:Biizero}).

Finally, it should be verified what happens, when the denominators of the expressions in Eqs.~(\ref{eq:B11zero}--\ref{eq:B33zero}) are zeroed. This basically occurs when from the reference configuration, a leg is turned at $90^\circ$ (in either direction), and possibly another one is turned at $180^\circ$. It can be verified that at such a configuration, a Type~1 singularity occurs. Indeed, the numerator of the above three equations also becomes zero.

In conclusion, the \textit{Agile Eye} has the following three families of singular configurations: 

\begin{itemize}
	\item \textbf{Six self motions} of the mobile platform described by Eqs.~(\ref{eq:selfmotion1}--f), in SO(3), and by Eqs.~(\ref{eq:cond1}--c), in the active-joint space. For each self motion, one of the legs is singular. For the active-joints corresponding to each of Eqs.~(\ref{eq:cond1}--c), the four (nontrivial) assembly modes degenerate in pairs into the two corresponding self motions.
	
  \item \textbf{Infinitely many infinitesimal motions} of the mobile platform at the four trivial orientations when the active-joints satisfy Eq.~(\ref{eq:Azero}) but neither of Eqs.~(\ref{eq:cond1}--c). Each of the four trivial orientations belongs to the sets of orientations corresponding to three self motions. For the active joint variables corresponding to these singular configurations, the only solutions to the direct kinematic problem are the four trivial ones.

  \item \textbf{Lockup configuration} described in SO(3) by each of the four trivial orientations and in the active-joint space space by anything but Eq.~(\ref{eq:Azero}). At these Type~1 singular configurations, the mobile platform cannot move even under external force. For the active joint variables corresponding to these singular configurations, the direct kinematic problem has eight solutions.
\end{itemize}

%%%%% SECTION 6 -----------------------------------------------------------
\section{Working Modes and Assembly Modes}

As was mentioned before, a particularity of the \textit{Agile Eye} is that the determinant of matrix $\negr A$, Eq.~(\ref{eq:Azero}), is a function of the active-joint variables only and has the same value for all four nontrivial assembly modes. Equation~(\ref{eq:Azero}) represents a surface that divides the active-joint space into two domains where $\det(\negr A)$ is either positive or negative. A connectivity analysis was made on these two domains to prove this property.

The images of these two domains in the workspace yield eight identical domains (the whole orientation space without the singularity curves), each one being associated with one of the eight working modes. It can been seen geometrically, or proved algebraically, that each of the four nontrivial assembly modes correspond to a different working mode. Therefore, changing an assembly mode inevitably requires a singularity to be crossed. Hence, the \textit{Agile Eye} is not cuspidal, meaning that as long as it does not cross a singularity, it remains in a single working mode and in a single assembly mode.

The eight working modes are divided into two groups. Depending on the sign of $\det(\negr A)$, the four assembly modes each correspond to a working mode from one of these groups. In Table~\ref{table_solution_DKP_pii}, it is assumed that for the first solution of the direct kinematic problem the sign of $B_{ii}$ given in Eqs.~(\ref{eq:B11zero}--\ref{eq:B33zero}) is negative. For the other solutions, $\{\psi, \theta, \phi\}$ are replaced by the value given in Table~\ref{table_solution_DKP} and the influence on the sign of $B_{ii}$ is shown using simple trigonometric properties. Recall that $|B_{ii}|$ does not change when an assembly mode is changed.

%%%%% TABLE 2 %%%%%%%%----------------------------------------------------------
\begin{table}[t]
   \center
   \begin{tabular}{|c|c|c|c|c|} \hline
     Sign of & $B_{11}$ & $B_{22}$ & $B_{33}$ & $\det(\negr B)$\\ \hline
     Solution 1 & $-$ & $-$ & $-$ & $-$\\
     Solution 2 & $+$ & $+$ & $-$ & $-$\\
     Solution 3 & $-$ & $+$ & $+$ & $-$\\
     Solution 4 & $+$ & $-$ & $+$ & $-$\\ \hline
   \end{tabular}
   \caption{The signs of $\negr B_{ii}$ for a given set of active-joint variables}
   \label{table_solution_DKP_pii}
\end{table}
%%%%%%%%%%%%%%%%%%%%%%----------------------------------------------------------

\begin{figure}[t]
  \begin{center}
    {\includegraphics[width=42mm]{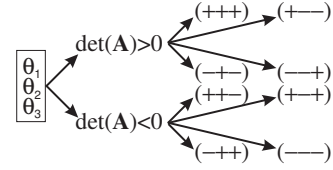}}
    \caption{The eight assembly modes characterized by the working mode}
    \protect\label{fig:working_assembly_mode}
  \end{center}
\end{figure}

Figure~\ref{fig:working_assembly_mode} summarizes the behavior of the direct kinematic problem where a solution can be chosen according to the sign of $\negr B_{ii}=(+++)$ for example and the sign of $\det(\negr A)$. Similarly, when the inverse kinematic model is solved, the working mode is easily characterized by the sign of $\negr B_{ii}$.

%%%%% FIGURE 3 %%%%%%%%---------------------------------------------------------
%%%%%%%%%%%%%%%%%%%%%%%---------------------------------------------------------

%%%%% SECTION 7 -----------------------------------------------------------
\section{Conclusions}

An in-depth kinematic analysis of a special spherical parallel wrist, called the \textit{Agile Eye}, was done, pinpointing some important facts that were previously overlooked. It was demonstrated that  the workspace of the \textit{Agile Eye} is unlimited and flawed only by six singularity curves (rather than surfaces). Furthermore, these curves were shown to correspond to self-motions of the mobile platform. It was also proved that the four assembly modes of the \textit{Agile Eye} are directly related to the eight working modes and the sign of the determinant of one of the Jacobian matrices. It was shown that as long as the \textit{Agile Eye} does not cross its singularity curves, it remains in a single working mode and in a single assembly mode.

%%%%% SECTION 8 -----------------------------------------------------------
%\section*{Acknowledgments}

%This work was supported by the \textit{Fonds québécois de la recherche sur la nature et les technologies} and was done during a two-month visit of the second author at the \'Ecole de Technologie Sup\'erieure.

%%%%% BIBLIOGRAPHY --------------------------------------------------------
\bibliographystyle{unsrt}

\end{document}